\documentclass[pdflatex,sn-mathphys-num]{sn-jnl}

\usepackage{graphicx}
\usepackage{multirow}
\usepackage{amsmath,amssymb,amsfonts}
\usepackage{booktabs}
\usepackage{algorithm}
\usepackage{algorithmicx}
\usepackage{algpseudocode}

\begin{document}

\title[CoFreeVLA]{CoFreeVLA: Short-Horizon Collision-Free Dual-Arm Manipulation via Vision-Language-Action Model and Risk Estimation}

\author[1]{\fnm{Yaohua} \sur{Liu}}\email{liuyaohua@hqcmlab.cn}
\author[2]{\fnm{Binkai} \sur{Ou}}\email{yoanmike666@gmail.com}
\author*[3]{\fnm{Hengjun} \sur{Zhang}}\email{zhanghengjun@guet.edu.cn}

\affil[1]{\orgname{Chinese Medicine Guangdong Laboratory}, \orgaddress{\street{Hengqin}, \city{Zhuhai}, \postcode{519031}, \state{Guangdong}, \country{China}}}
\affil[2]{\orgname{BoardWare Information System Company}, \orgaddress{\city{Macau}, \postcode{999078}, \country{China}}}
\affil[3]{\orgdiv{School of Electronic Engineering and Automation}, \orgname{Guilin University of Electronic Technology}, \orgaddress{\city{Guilin}, \postcode{541000}, \state{Guangxi}, \country{China}}}

\abstract{Vision–Language–Action (VLA) models enable instruction-following manipulation, yet their deployment on coordinated dual-arm platforms remains severely constrained by under-modeled self-collisions between manipulators and grasped objects. To address this critical safety gap, we propose CoFreeVLA, a novel framework that augments end-to-end VLA policies with a lightweight, short-horizon self-collision risk estimator. The estimator predicts collision likelihoods directly from proprioceptive states, visual embeddings, and candidate action sequences. Deeply integrated into the closed-loop control system, this estimator proactively gates risky commands, autonomously synthesizes recovery trajectories to safe states via risk-guided adjustments, and biases policy refinement for safer rollouts. To ensure robust calibration, the estimator utilizes a two-stage training pipeline: pre-training with model-based synthetic collision labels, followed by post-training on real-robot rollouts. Across five bimanual tasks, six VLA backbones, and 30 trials per variant, the task-averaged collision rate decreases from 0.54 to 0.23, while the task-averaged success rate increases from 0.45 to 0.61. Compared to representative baselines, CoFreeVLA substantially reduces self-collision frequencies and improves overall task success rates, providing a crucial step toward the safe deployment of foundational models in multi-arm continuous control.}

\maketitle

\section{Introduction} \label{Sec: intro}

Vision–Language–Action (VLA) models have recently emerged as a powerful paradigm for robotic control, demonstrating remarkable capabilities in grounding natural language instructions into visual observations and executable actions~\cite{brohan2024rt,driess2023palm,team2024octo,kim2024openvla,ahn2022can}. While these models have shown impressive generalization in single-arm manipulation tasks, extending them to dual-arm systems introduces new challenges. Dual-arm manipulation requires coordination across two high-degree-of-freedom manipulators, making spatial reasoning and safety-critical planning considerably more complex~\cite{abbas2023systematic, bi2025vla, zhai2025vfp}. Existing approaches largely focus on avoiding external collisions with the environment or task objects—via classical trajectory optimization and widely deployed toolchains—yet self-collisions between the two arms or between an arm and a grasped object remain underexplored in end-to-end VLA settings~\cite{zucker2013chomp,schulman2014motion,moveit_ompl_collision_noetic,moveit_planning_scene_self_collision,pan2012fcl,lei2020real,fresnillo2023extending}. Such self-collisions can result in task failure, hardware damage, or unsafe behaviors, and therefore pose a fundamental obstacle to deploying VLA-based dual-arm systems in real-world environments.

Classical motion planners and engineering toolchains have been widely used to avoid collisions with the environment and with task objects \cite{zucker2013chomp,schulman2014motion,pan2012fcl,lei2020real,fresnillo2023extending}. Those approaches are effective for planning geometry-aware trajectories but typically operate in separate modules from learned VLA policies, which limits the responsiveness and end-to-end adaptability of learned controllers. In particular, self-collisions between the two arms and collisions involving a grasped object are often overlooked in end-to-end VLA pipelines despite being frequent causes of task interruption, hardware stress, and unsafe behavior. This gap is especially consequential for real-world dual-arm deployment where perceptual uncertainty, kinematic calibration errors, and dynamic interactions with objects raise the risk of unintended contact.

To address this gap, we propose CoFreeVLA, a risk-aware VLA framework designed for collision-free dual-arm manipulation. At the core of CoFreeVLA is a self-collision risk estimator that predicts the likelihood of future collisions given the current observation and planned actions from the VLA policy. This risk estimator is integrated into the control loop in three complementary ways: (i) by halting unsafe actions before collisions occur, (ii) by guiding the system to recover toward safe configurations when risk is detected, and (iii) by refining the VLA policy itself through optimization with risk-aware feedback. By embedding explicit collision reasoning into VLA decision-making—complementary to classical safety layers such as control barrier functions~\cite{ames2016control,ames2019control}—CoFreeVLA enables safer and more reliable dual-arm manipulation without sacrificing task performance.

We evaluate CoFreeVLA across a set of dual-arm manipulation tasks that highlight the need for safe spatial reasoning, including handover, cooperative grasping, and assembly~\cite{grotz2024peract2,fu2024mobile, yu2024manipose}. To train and validate the risk estimator, we employ a two-stage strategy: pre-training with synthetic collision data generated from a model-based checker~\cite{pan2012fcl,moveit_ompl_collision_noetic,moveit_planning_scene_self_collision}, followed by post-training with real-robot rollouts under VLA policies. Our experiments assess both the predictive accuracy of the risk estimator and the downstream benefits of risk-aware policy refinement. We benchmark CoFreeVLA against recent dual-arm learning baselines, including APEX and RDT-1B~\cite{dastider2024apex,liu2024rdt}, on an AgileX-based dual-arm platform~\cite{agilex_global}, measuring performance in terms of task success, collision rate, and recovery effectiveness.

In summary, our contributions are as follows:
\begin{enumerate}
    \item We propose CoFreeVLA, a framework that addresses the overlooked problem of inter-arm and arm–object collision in dual-arm VLA-based manipulation.
    \item We design a self-collision risk estimator that predicts collision likelihood from visual observations and action sequences, and integrate it into execution, recovery, and policy refinement.
    \item We conduct extensive experiments on dual-arm manipulation tasks, validating the effectiveness of CoFreeVLA in reducing risky collisions while maintaining task performance.
\end{enumerate}

The remainder of this paper is organized as follows. Section~\ref{Sec: related work} reviews VLA models and collision avoidance. Section~\ref{Sec: methodology} formulates the control problem and presents CoFreeVLA, including estimation, recovery, and training. Section~\ref{Sec: experiment} evaluates transferability, task-level safety, ablations, and real-time control. Section~\ref{Sec: conclusion} summarizes the findings and discusses future directions.

\section{Related Work} \label{Sec: related work}

\subsection{Vision--Language--Action Models}
Vision--language--action models combine visual--linguistic representations with learned control policies to map natural-language instructions and sensory observations to robot actions. Early systems established this paradigm by transferring large-scale multimodal knowledge to embodied control. RT-2 connected web-scale vision--language pretraining to end-to-end robotic actions, while PaLM-E incorporated visual, linguistic, and proprioceptive inputs for embodied reasoning \cite{brohan2024rt,driess2023palm}. Octo and OpenVLA subsequently scaled generalist policy learning through heterogeneous robot data and open-source pretraining, respectively \cite{team2024octo,kim2024openvla}. Compact models such as TinyVLA target lower inference cost and improved data efficiency through lightweight multimodal backbones and diffusion-based action decoding \cite{wen2025tinyvla}.

$\pi$0.5 extends this direction through co-training on heterogeneous multi-robot and multimodal data, with high-level semantic supervision supporting long-horizon manipulation \cite{intelligence2025pi05}. SmolVLA focuses on accessible deployment through asynchronous inference and chunked action generation on consumer hardware \cite{shukor2025smolvla}. Other recent efforts address cross-embodiment transfer and decoding efficiency: X-VLA uses embodiment-specific soft prompts, PD-VLA accelerates action-chunk decoding through parallel fixed-point iterations, and UniVLA learns task-centric latent actions from heterogeneous videos \cite{zheng2025x,song2025accelerating,bu2025univla}. SayCan further influenced VLA design by using language-model affordances for skill selection \cite{ahn2022can}. Despite these advances in instruction following, generalization, and efficiency, existing VLA policies typically rely on data priors or external planning modules for safety and do not explicitly forecast imminent self-collision from perceptual inputs and candidate dual-arm actions. CoFreeVLA addresses this gap by coupling a compact risk predictor with execution-time gating and recovery.

\subsection{Collision Avoidance for Robot Manipulation}
Collision avoidance in manipulation spans geometric planning, collision detection, feedback control, and learned motion generation. Functional-gradient planning and sequential convex optimization incorporate distance costs into trajectory generation, while GPU-oriented toolchains accelerate parallel collision queries and constrained replanning \cite{zucker2013chomp,schulman2014motion,sundaralingam2023curobo,pan2012fcl}. Robot-collision surveys further emphasize detection under sensing and model uncertainty rather than geometric post-processing alone \cite{haddadin2017robot}. This issue is particularly important for dual-arm systems, where links, grippers, and grasped objects can collide during otherwise task-valid motions.

At the control level, control-barrier-function filters provide formal set-invariance guarantees, and collision-free model-predictive control extends safety constraints to whole-body motion \cite{ames2016control,ames2019control,chiu2022collision}. Dual-arm systems have been addressed through kinematic avoidance rules, MoveIt extensions, neural self-collision controllers, and spatiotemporal diffusion policies \cite{lei2020real,fresnillo2023extending,kramar2022self,lv2025spatial}. Other learning-based approaches combine collision-aware representations with bimanual action generation, including VLM-assisted flow-diffusion, manipulability-aware diffusion, and collision-free latent motion generation \cite{chen2025vlm,li2025manidp,dastider2024apex}. Learned collision detectors can complement geometric checkers in cluttered scenes \cite{golluccio2025deep,zhu2025efficient}. However, these methods typically plan in geometric state spaces, intervene at the low-level controller, or generate collision-aware actions without forecasting imminent self-collision from perceptual inputs and candidate VLA actions. CoFreeVLA addresses this gap by coupling a compact risk predictor with execution gating, recovery, and policy refinement.

\section{Methodology} \label{Sec: methodology}

\subsection{Problem formulation}

We consider a dual-arm manipulator whose left and right arms have
$n_L$ and $n_R$ degrees of freedom, respectively. At time $t$, its joint
configuration is
\begin{equation}
q_t = \big(q_t^{L}, q_t^{R}\big)\in\mathbb{R}^{n_L+n_R},
\label{eq:joint}
\end{equation}
subject to the prescribed joint-position and joint-velocity limits. For
each arm $k\in\{L,R\}$, let $x_t^k\in SE(3)$ denote the end-effector pose
and $g_t^k$ the gripper state. The action consists of Cartesian
increments of the two end-effectors,
\begin{equation}
 \Delta x_t = \big(\Delta x_t^{L}, \Delta x_t^{R}\big),
\label{eq:action}
\end{equation}
which we identify with the control command as $a_t \equiv \Delta x_t.$

The decision state is defined as
\begin{equation}
s_t = \big(x_t, g_t, o_t\big),
\label{eq:state}
\end{equation}
where $x_t$ and $g_t$ collect the end-effector poses and gripper states
of both arms, respectively, and $o_t$ comprises onboard perceptual
observations, such as RGB-D data, together with the task instruction.
An end-to-end vision--language--action policy maps images, language, and
proprioceptive measurements to executable actions. Its stochastic form
is expressed as
\begin{equation}
a_t \sim\pi_{\theta}(\cdot\mid s_t),
\label{eq:policy_stochastic}
\end{equation}
or, equivalently, its deterministic implementation can be written as
\begin{equation}
a_t = f_{\theta}(s_t).
\label{eq:policy_deterministic}
\end{equation}

\subsection{Self-collision risk estimator}

\begin{figure*}[t]
    \centering
    \includegraphics[width=0.8\linewidth]{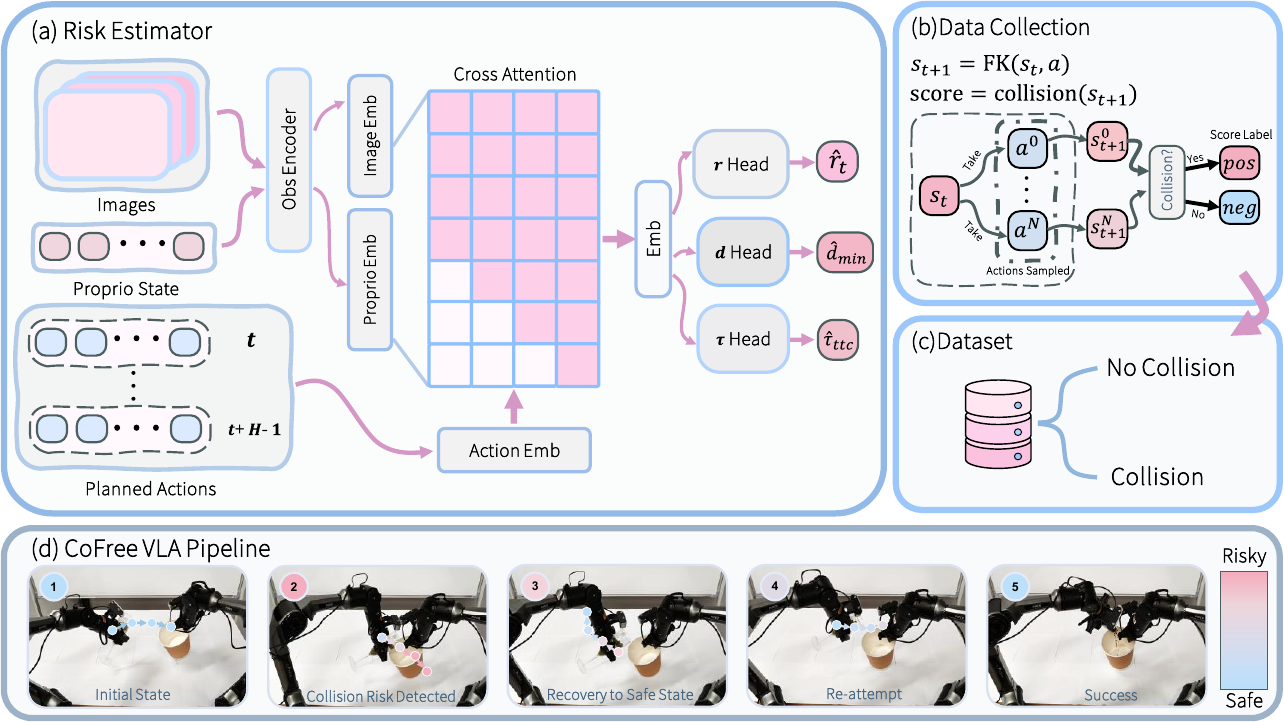}
    \caption{\textbf{(a) Risk Estimator.} Given a state–action pair, the model applies a cross-attention module followed by three heads to predict collision risk $r$, minimum distance $d$, and time-to-collision $\tau$. \textbf{(b) Data Collection} and \textbf{(c) Dataset.} For each state, we sample candidate actions and check their outcomes for labeling.
\textbf{(d) CoFree VLA.} At run time, the system continuously estimates risk; when it exceeds a threshold, the policy will be blocked and the arms will be returned to a safe configuration, after which the policy resumes to complete the task.}
    \label{Fig: method}
\vspace{-4mm}
\end{figure*}

As in Fig.~\ref{Fig: method}(a), the estimator takes the current dual-arm state and a short-horizon plan as input. We form the action sequence $A_{t:t+H-1}$ from the VLA policy, pair it with proprioceptive state $(x_t,g_t)$, and a visual embedding $z_t=f_{\text{VLA}}(o_t)$ from the VLA backbone. The estimator predicts a calibrated self-collision risk $\hat r_t$ and optionally geometric surrogates such as the minimum inter-body distance $\hat d_{\min}$ (mm) and time-to-collision $\hat\tau_{\text{ttc}}$ (s), produced by auxiliary heads:
\begin{align}
\hat r_t &= E_{\phi}\big(x_t, g_t, z_t, A_{t:t+H-1}\big) \in [0,1],
\label{eq:risk}\\
\hat d_{\min,t} &= D_{\phi}\big(x_t, g_t, z_t, A_{t:t+H-1}\big)\in\mathbb{R}_{\ge 0},
\label{eq:dmin}\\
\hat\tau_{\mathrm{ttc},t} &= T_{\phi}\big(x_t, g_t, z_t, A_{t:t+H-1}\big)\in\mathbb{R}_{\ge 0}.
\label{eq:ttc}
\end{align}

The model architecture encodes inputs as two parallel streams: (i) an observation stream and (ii) an action stream. The observation stream encodes a compact feature $z_t$ taken from the penultimate VLA image token together with the proprio state $(x_t,g_t)$; the action stream encodes $A_{t:t+H-1}$ as a length-$H$ token sequence with positional encodings.  A lightweight temporal encoder (MLP for small $H$, or a small Transformer for $H\!\ge\!5$) fuses the two streams via cross-attention, followed by pooling and prediction heads for $\hat r_t$, $\hat d_{\min}$, and $\hat\tau_{\text{ttc}}$. We apply temperature scaling to calibrate $\hat r_t$, and size the network for sub-5\,ms inference so it can run at control rates (10–50\,Hz).

As illustrated in Fig.~\ref{Fig: method}(b) and (c), we construct a dataset
\begin{equation}
\mathcal{D}
=
\left\{
\left(x_t,g_t,z_t,A_{t:t+H-1},y_t\right)
\right\}_{t=1}^{N},
\label{eq:d}
\end{equation}
where \(y_t=\{y_t^{\mathrm{bin}},y_t^{d},y_t^{\mathrm{ttc}}\}\) denotes the corresponding labels. During pre-training, we use a model-based checker, such as a mesh-distance checker, to evaluate candidate action sequences \(A_{t:t+H-1}\) generated by the VLA with added jitter. For each sequence, we record a binary indicator \(y_t^{\mathrm{bin}}\in\{0,1\}\) of whether a collision occurs within the horizon \(H\), the minimum inter-arm or arm--object distance \(y_t^{d}\), and the first time to collision \(y_t^{\mathrm{ttc}}\). We emphasize near-collision cases by oversampling samples with small \(y_t^{d}\). During post-training, we relabel real-robot rollouts using contact events, torque spikes, and visual overlap to account for calibration and grasp uncertainties.

\subsection{Collision-Free VLA Structure}
As summarized in Algorithm~\ref{alg:main:extended}, Collision-Free VLA operates in three stages: 1) risk estimation to stop unsafe actions, 2) recovery to safe states, and 3) policy optimization for collision-free control.

\textbf{Risk estimation to stop.}
In Fig.~\ref{Fig: method}(d), at each control step, the VLA proposes a short-horizon plan $A_{t:t+H-1}=(a_t,\ldots,a_{t+H-1})$. We evaluate the self-collision risk $\hat r_t=E_\phi(s_t,A_{t:t+H-1})\in[0,1]$. If $\hat r_t>\tau_\uparrow$, execution of $a_t$ is blocked and the controller enters a protective mode; execution resumes only when $\hat r_t\le\tau_\downarrow$ (hysteresis with $\tau_\downarrow<\tau_\uparrow$ reduces chattering). As an optional soft gate when not blocked, commanded velocities are scaled by $\alpha_t=\mathrm{clip}(1-\hat r_t/\tau_\uparrow,0,1)$ before sending to the low-level controller.

\textbf{Recovery to safe states.}
When blocked, we synthesize a short recovery sequence $A^{\mathrm{rec}}_{t:t+h-1}$ that drives the system toward the self-safe set $\mathcal{S}_{\mathrm{safe}}=\{\,q:\,d_{\min}(q)\ge d_0\,\}$. We optimize a risk-shaped objective $A^{\mathrm{rec}}_{t:t+h-1}\leftarrow\arg\min_{A}\;E_\phi(s_t,A)+\lambda\|A\|^2$ under joint and velocity limits, solved by a few projected-gradient steps. This locally reduces predicted collision risk; recovery terminates once $\hat r_t\le\tau_\downarrow$ for $K$ consecutive cycles.

\textbf{Policy optimization for collision-free control.}
Beyond online gating, we fine-tune the VLA with a safety-aware objective that trades off task intent against predicted risk. We minimize a surrogate that keeps the refined plan close to the original while lowering $E_\phi$: $\min_{\theta}\ \mathbb{E}_{s_t}\big[\alpha\,\|A'_{t:t+H-1}-A_{t:t+H-1}\|^2+\beta\,E_\phi(s_t,A'_{t:t+H-1})\big]$, where $A'_{t:t+H-1}$ is obtained by rolling out $f_\theta(s_t)$ for $H$ steps. Practically, we perform (i) supervised fine-tuning on a filtered dataset $\mathcal{D}_{\mathrm{safe}}=\{(s,a):E_\phi(s,A_{t:t+H-1})\le\tau_\downarrow\}$ with risk-weighted losses $w(s)=\exp(-\kappa\,E_\phi(s,A_{t:t+H-1}))$, biasing the model toward a collision-free target distribution; and (ii) test-time action refinement by solving $\min_{A'}\ \alpha\,\|A'-A_{t:t+H-1}\|^2+\beta\,E_\phi(s_t,A')$ under control limits, then executing only the first optimized step to preserve reactivity. This balances task success with collision-free behavior.

\subsection{Training Procedure for CoFreeVLA}
\textbf{Risk estimator pre-training.}
We first pre-train $E_\phi$ offline using model-based collision labels. Given a state $s_t$, we generate horizon plans $A_{t:t+H-1}$ by rolling out the base VLA with beam variants and added jitter. We collect these plans in simulation and on the real robot operating in guard mode. For each pair $(s_t,A_{t:t+H-1})$, mesh-distance queries and a predictive checker provide three labels. The first is a binary indicator $y_t^{\mathrm{bin}}\in\{0,1\}$ of collision within $H$. The second is the minimum inter-arm or arm--grasped-object distance $y_t^{d}$ (mm) over the horizon. The third is the first time to collision $y_t^{\mathrm{ttc}}$ (s). The training objective combines binary cross-entropy (BCE) and regression terms:
\begin{equation}
\begin{aligned}
\mathcal{L}={}&\lambda_{\mathrm{bce}}\mathrm{BCE}(\hat r_t,y_t^{\mathrm{bin}})
+\lambda_d\|\hat d_{\min}-y_t^{d}\|_2^2 +\lambda_{\mathrm{ttc}}\|\hat\tau_{\mathrm{ttc}}-y_t^{\mathrm{ttc}}\|_1.
\end{aligned}
\label{eq:loss}
\end{equation}
We address class imbalance by assigning higher weights to false negatives and oversampling near-miss samples with small $y_t^{d}$. We also apply a horizon curriculum that begins with a short $H$ and gradually increases it. After training, temperature scaling on a held-out set calibrates $\hat r_t\in[0,1]$.

\textbf{Closed-loop fine-tuning.}
We then adapt both $E_\phi$ and the VLA through safety-supervised closed-loop fine-tuning. With the risk gate $(\tau_\uparrow,\tau_\downarrow)$ enabled, we collect tuples $(s_t,A_{t:t+H-1},\hat r_t)$ together with contact events and recovery-corrected actions. These data are aggregated in a DAgger-like buffer and support two updates. First, we post-train $E_\phi$ on real-robot labels to account for calibration and grasp uncertainty. Second, we refine the VLA using the risk-weighted objective
\begin{equation}
\min_{\theta}\ \mathbb{E}_{(s,a)\sim\mathcal{D}}\!\left[
\ell_{\mathrm{BC}}(\pi_\theta(s),a)\,w(s)\right],
\label{eq:bc_refine}
\end{equation}
where $w(s)=\exp\!\big[-\kappa E_\phi(s,A_{t:t+H-1})\big]$. When available, pairwise preferences between safe and unsafe variants or a small reinforcement-learning term can provide an additional risk penalty. We tune the thresholds through receiver operating characteristic (ROC) analysis while prioritizing a low false-negative rate. Hysteresis suppresses chattering, and a watchdog halts execution if $\hat r_t$ saturates.
\begin{algorithm}
\caption{Collision-Free VLA Execution, Recovery, and Policy Refinement}
\label{alg:main:extended}
\begin{algorithmic}[1]
\Require policy \(f_{\theta}\), estimator \(E_{\phi}\), buffer \(\mathcal{B}\), thresholds \(\tau_{\downarrow}<\tau_{\uparrow}\), horizons \(H,h\), and termination window \(K\)
\State observe \(s_t=(x_t,g_t,o_t)\) and compute \(z_t=f_{\mathrm{VLA}}(o_t)\)
\State generate \(N\) candidate plans \(\{A^{(n)}_{t:t+H-1}\}_{n=1}^{N}\) from \(f_{\theta}(s_t)\) with beam search and jitter
\For{each candidate \(A^{(n)}_{t:t+H-1}\)}
    \State evaluate \(\hat r_t^{(n)}\) and, optionally, \(\hat d_{\min,t}^{(n)}\) and \(\hat\tau_{\mathrm{ttc},t}^{(n)}\)
\EndFor
\State \(n^*\leftarrow\arg\min_{n\in\{1,\ldots,N\}}\hat r_t^{(n)}\)
\State \(A^*_{t:t+H-1}\leftarrow A^{(n^*)}_{t:t+H-1}\), \(\hat r_t^*\leftarrow\hat r_t^{(n^*)}\)
\If{\(E_{\phi}\) is differentiable and the real-time budget permits}
    \State refine \(A^*_{t:t+H-1}\) by minimizing its deviation from the proposal and its predicted risk
    \State reevaluate \(\hat r_t^*\), \(\hat d_{\min,t}^*\), and \(\hat\tau_{\mathrm{ttc},t}^*\)
\EndIf
\If{\(\hat r_t^*\le\tau_{\downarrow}\)}
    \State execute the first action \(a_t^*\) of \(A^*_{t:t+H-1}\); set \(a_t^{\mathrm{exec}}\leftarrow a_t^*\)
\ElsIf{\(\hat r_t^*>\tau_{\uparrow}\)}
    \State solve \(A^{\mathrm{rec}}_{t:t+h-1}\leftarrow\arg\min_A E_{\phi}(s_t,A)+\lambda\lVert A\rVert^2\) under the control limits
    \State execute its first action \(a_t^{\mathrm{rec}}\); set \(a_t^{\mathrm{exec}}\leftarrow a_t^{\mathrm{rec}}\)
    \State keep recovery mode active until \(\hat r_t\le\tau_{\downarrow}\) for \(K\) consecutive cycles
\Else
    \State \(\alpha_t\leftarrow\mathrm{clip}(1-\hat r_t^*/\tau_{\uparrow},0,1)\)
    \State execute \(a_t^{\mathrm{exec}}\leftarrow\alpha_t a_t^*\)
\EndIf
\State append \((s_t,A^*_{t:t+H-1},a_t^{\mathrm{exec}},\hat r_t^*,\hat d_{\min,t}^*,\hat\tau_{\mathrm{ttc},t}^*)\) to \(\mathcal{B}\)
\If{a post-training update is scheduled}
    \State update \(E_{\phi}\) using the composite loss in Eq.~\eqref{eq:loss}
    \State update \(f_{\theta}\) using the risk-weighted objective in Eq.~\eqref{eq:bc_refine}
\EndIf
\end{algorithmic}
\end{algorithm}

\subsection{Implementation Notes}
We use a PyBullet 3.2.6 digital twin with an EGL RGB-D renderer and the Flexible Collision Library (FCL), version 0.7.0 \cite{pan2012fcl}, for nominal pre-training. PyBullet advances the AgileX-based PIPER-style URDF scene and renders synchronized observations. FCL independently processes the corresponding inflated link and object meshes to generate collision and proximity labels. For simulation pre-training, robot-link and gripper meshes are loaded from the PIPER-style URDF package. Environment obstacles and task objects are represented by CAD meshes or dimension-matched primitives in the randomized digital twin.

Each sample contains a scene, a joint state, a rendered RGB-D observation, and a candidate action chunk. We generate 200,000 samples and partition complete scene trajectories into 80\% training, 10\% calibration, and 10\% test splits. This trajectory-level partition prevents adjacent chunks from crossing splits. Candidate chunks comprise nominal policy actions, bounded perturbations that represent proposal and tracking variation, and boundary-seeking actions that increase coverage near small clearances. Cartesian increments are converted to joint trajectories using warm-started damped-least-squares inverse kinematics. Chunks without a joint-limit-feasible solution are marked infeasible and never receive a safe label.

For hardware adaptation, robot, tool, and task-object geometries are represented by calibrated copies of the corresponding URDF, CAD, or dimension-measured models. These models are aligned with the physical workspace through the robot and camera calibration transforms. Real RGB-D frames are consumed by the estimator and FCL replays measured joint and object poses against the calibrated geometries to provide model-based clearance and $\hat\tau_{\text{ttc}}$  targets. Hardware adaptation initializes the estimator from the nominal checkpoint and fine-tunes it on the 350 training rollouts for at most 30 epochs. The RGB-D encoder is frozen for the first three epochs to stabilize the geometric heads, after which all estimator layers are jointly updated. Training windows are sampled by rollout, with at most one window per control cycle; no window from the calibration or test partitions is used for optimization. The calibration split is used first for temperature scaling and then for selecting the frozen risk and $\hat\tau_{\text{ttc}}$  operating points. Once these quantities are fixed, all estimator parameters and calibration values remain unchanged during closed-loop evaluation.

As illustrated in Fig.~\ref{Fig: method}(a), the risk estimator adopts a four-layer temporal Transformer backbone with a model width of 384 and eight attention heads. Prior to multimodal fusion, the observation and action streams are projected into this shared latent dimension. Each Transformer layer is equipped with pre-layer normalization, a gated feed-forward block, and a dropout rate of 0.1. The proprioceptive measurements together with $H$ standardized future actions are encoded as temporal tokens, which are further conditioned on the visual embedding $z_t$ via cross-attention operations. The learned temporal representations are then shared across three task-specific prediction heads. Among them, the $\hat d_{\min}$ head and the $\hat{\tau}_{\text{ttc}}$ head regress continuous physical quantities around collision boundaries, addressing the limitation that binary collision labels fail to characterize relative spatial proximity.

All training runs use a single NVIDIA GeForce RTX 4090 GPU with 24\,GB of memory. Synthetic pre-training uses AdamW with a batch size of 512 and a learning rate of $1\times10^{-4}$, whereas hardware adaptation uses batch size 64 and the same learning rate. In Eq.~\eqref{eq:loss}, we set $\lambda_{\mathrm{bce}}=1.0$, $\lambda_d=0.5$, and $\lambda_{\mathrm{ttc}}=0.1$. These weights balance binary collision classification against the two auxiliary geometric regression objectives. Pre-training takes approximately 3\,h, whereas real-robot post-training takes approximately 45\,min.

At deployment, we run the VLA~\cite{liu2024rdt} at 10\,Hz and the low-level controller at 30\,Hz. The deployed action horizon is $H=16$. Using the calibration split described above, we select the risk thresholds, risk weight $\kappa$, and recovery penalty $\lambda$. We set $\tau_\uparrow=0.85$ to trigger blocking and $\tau_\downarrow=0.30$ to resume execution. This hysteresis reduces oscillatory switching during precision tasks. At each VLA update, $E_\phi$ evaluates $N$ beam-and-jitter candidates, and the controller executes the first action of the lowest-risk feasible sequence. We export $E_\phi$ to ONNX or TorchScript and prefetch $z_t$ from the VLA backbone. These optimizations maintain a sub-5\,ms forward pass on a CPU or GPU. When gradient-based refinement is unavailable, recovery uses a quadratic-programming/control-barrier-function (QP/CBF) fallback with constraints derived from the distance head.

\section{Experiment} \label{Sec: experiment}
\begin{figure*}[t]
    \centering
    \includegraphics[width=0.95\linewidth]{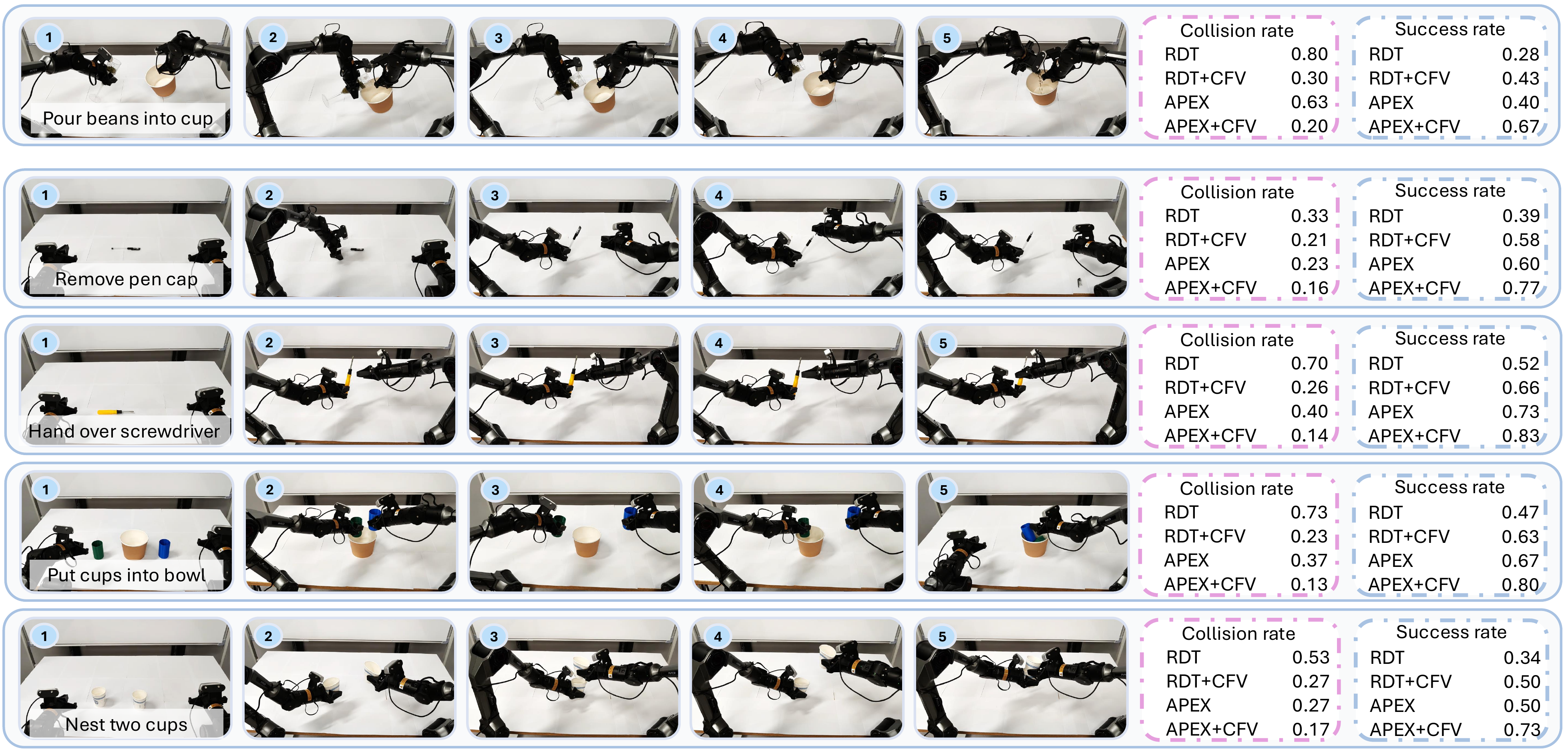}
    \caption{Experimental results in 5 dual-arm manipulation tasks. Evaluations of collision and success rates are on 30 trials}
    \label{fig:exp}
\vspace{-4mm}
\end{figure*}
\subsection{Experiment Setup}
All experiments are conducted on an AgileX-based PIPER-style bimanual mobile manipulator. To systematically evaluate the system, we select five representative bimanual tasks that span a wide spectrum of inter-arm interference and precision demands:
\begin{itemize}
    \item \textbf{Pour beans into cup:} High spatial overlap with a high risk of inter-arm and arm-object collisions.
    \item \textbf{Hand over screwdriver:} Requires close-proximity coordination and dynamic object transfer.
    \item \textbf{Put cups into bowl (Tubes):} Emphasizes spatial reasoning under restricted clearance.
    \item \textbf{Remove pen cap \& Nest two cups:} High-precision tasks demanding millimeter-level alignment where arms must naturally operate in extreme proximity.
\end{itemize}

We employ six structurally diverse baseline policies as the underlying
action policies: two lightweight models, TinyVLA~\cite{wen2025tinyvla}
and SmolVLA~\cite{shukor2025smolvla}; three representative
vision--language--action (VLA) models, OpenVLA~\cite{kim2024openvla},
RDT-1B~\cite{liu2024rdt}, and $\pi$0.6~\cite{intelligence2025pi}; and a
dual-arm-specific generative policy, APEX~\cite{dastider2024apex}. For
each task and method variant, we conduct 30 trials. We report two
primary metrics: the Collision Rate (CR), defined as the proportion of
trials involving self-collision, and the Success Rate (SR), defined as
the proportion of trials satisfying the task-completion criteria.
Aggregated results are presented in Fig.~\ref{fig:exp}.

\subsection{Self-Collision Risk Estimation}
Table~\ref{tab:main_results_comprehensive} and Fig.~\ref{fig:exp} show that the short-horizon risk estimator and execution gate improve both safety and task completion for every evaluated backbone. Averaged across the six backbones and five tasks, the task-averaged CR decreases from 0.54 to 0.23, corresponding to an absolute reduction of 0.31 (57.6\%). Over the same trials, the task-averaged SR increases from 0.45 to 0.61, an absolute gain of 0.16 (34.5\%). The consistent direction of these changes across all policy families indicates that the execution-time safety layer is not dependent on a particular VLA architecture.

The backbone-level averages further quantify this consistency. CoFreeVLA reduces Avg. CR by 0.40 for TinyVLA, 0.35 for OpenVLA, 0.37 for RDT-1B, 0.28 for SmolVLA, 0.24 for $\pi$0.6, and 0.22 for APEX. The corresponding Avg. SR gains are 0.16, 0.16, 0.16, 0.15, 0.13, and 0.18, respectively. Thus, the framework provides measurable improvements for lightweight, foundation, and dual-arm-specific policies, while the magnitude of the gain varies with the nominal policy and task geometry.

Task-level results show that the benefit is largest when the planned motion creates substantial geometric interference. For Pouring Beans, vanilla policies yield CR values of 0.63--0.93, whereas CoFreeVLA reduces them to 0.20--0.32; SR correspondingly increases from 0.20--0.40 to 0.40--0.67. The remaining tasks also show lower CR and higher SR for every backbone, although the CR reductions are generally smaller in the tight-clearance precision tasks. This trend is consistent with the estimator providing the greatest advantage when inter-arm or arm--object overlap is the dominant failure mode. At the same time, fixed thresholds can remain conservative when successful manipulation requires intentional close proximity, which defines an operating boundary for the current calibration.

\begin{table*}[t]
\centering
\caption{Collision Rates (CR $\downarrow$) and Success Rates (SR $\uparrow$) Across Diverse VLA Backbones on 5 Bimanual Tasks}
\label{tab:main_results_comprehensive}
\resizebox{\textwidth}{!}{
\begin{tabular}{ll cccc cccc cccc cc}
\toprule
\multirow{2}{*}{\textbf{Backbone}} & \multirow{2}{*}{\textbf{Method Variant}} & \multicolumn{2}{c}{\textbf{Pouring Beans}} & \multicolumn{2}{c}{\textbf{Pen Cap Removal}} & \multicolumn{2}{c}{\textbf{Tool Handover}} & \multicolumn{2}{c}{\textbf{Tubes Placement}} & \multicolumn{2}{c}{\textbf{Cups Nesting}} & \multicolumn{2}{c}{\textbf{Average}} \\
\cmidrule(lr){3-4} \cmidrule(lr){5-6} \cmidrule(lr){7-8} \cmidrule(lr){9-10} \cmidrule(lr){11-12} \cmidrule(lr){13-14}
& & CR & SR & CR & SR & CR & SR & CR & SR & CR & SR & \textbf{CR} & \textbf{SR} \\
\midrule

\multirow{2}{*}{\textbf{TinyVLA}}
& Vanilla & 0.93 & 0.20 & 0.40 & 0.33 & 0.73 & 0.43 & 0.80 & 0.40 & 0.63 & 0.27 & 0.70 & 0.33 \\
& \textbf{+ CoFreeVLA} & 0.32 & 0.40 & 0.28 & 0.50 & 0.30 & 0.57 & 0.29 & 0.53 & 0.31 & 0.43 & \textbf{0.30} & \textbf{0.49} \\
\midrule

\multirow{2}{*}{\textbf{OpenVLA}}
& Vanilla & 0.87 & 0.27 & 0.37 & 0.40 & 0.60 & 0.53 & 0.67 & 0.47 & 0.47 & 0.33 & 0.60 & 0.40 \\
& \textbf{+ CoFreeVLA} & 0.30 & 0.43 & 0.24 & 0.57 & 0.27 & 0.67 & 0.20 & 0.63 & 0.22 & 0.50 & \textbf{0.25} & \textbf{0.56} \\
\midrule

\multirow{2}{*}{\textbf{RDT-1B}}
& Vanilla & 0.80 & 0.28 & 0.33 & 0.39 & 0.70 & 0.52 & 0.73 & 0.47 & 0.53 & 0.34 & 0.62 & 0.40 \\
& \textbf{+ CoFreeVLA} & 0.30 & 0.43 & 0.21 & 0.58 & 0.26 & 0.66 & 0.23 & 0.63 & 0.27 & 0.50 & \textbf{0.25} & \textbf{0.56} \\
\midrule

\multirow{2}{*}{\textbf{SmolVLA}}
& Vanilla & 0.73 & 0.30 & 0.29 & 0.47 & 0.50 & 0.60 & 0.57 & 0.53 & 0.40 & 0.40 & 0.50 & 0.46 \\
& \textbf{+ CoFreeVLA} & 0.27 & 0.47 & 0.19 & 0.63 & 0.21 & 0.70 & 0.20 & 0.67 & 0.23 & 0.57 & \textbf{0.22} & \textbf{0.61} \\
\midrule

\multirow{2}{*}{\textbf{$\pi$0.6}}
& Vanilla & 0.67 & 0.37 & 0.27 & 0.57 & 0.43 & 0.70 & 0.47 & 0.63 & 0.33 & 0.47 & 0.43 & 0.55 \\
& \textbf{+ CoFreeVLA} & 0.23 & 0.57 & 0.16 & 0.70 & 0.18 & 0.77 & 0.17 & 0.73 & 0.20 & 0.63 & \textbf{0.19} & \textbf{0.68} \\
\midrule

\multirow{2}{*}{\textbf{APEX}}
& Vanilla & 0.63 & 0.40 & 0.23 & 0.60 & 0.40 & 0.73 & 0.37 & 0.67 & 0.27 & 0.50 & 0.38 & 0.58 \\
& \textbf{+ CoFreeVLA} & 0.20 & 0.67 & 0.16 & 0.77 & 0.14 & 0.83 & 0.13 & 0.80 & 0.17 & 0.73 & \textbf{0.16} & \textbf{0.76} \\
\bottomrule
\end{tabular}
}
\vspace{1ex} \raggedright
\footnotesize{\textit{Note}: ``Vanilla'' denotes the original baseline policy without safety interventions. Bold values in the Average columns summarize the safety (CR reduction) and task-execution (SR increase) improvements obtained with our CoFreeVLA framework. The largest CR reductions occur on the interference-heavy Pouring Beans task, while precision tasks retain consistent reductions in CR and corresponding gains in SR.}
\end{table*}
\subsection{Collision-Free VLA}
\begin{figure*}[t]
    \centering
    \includegraphics[width=0.84\linewidth]{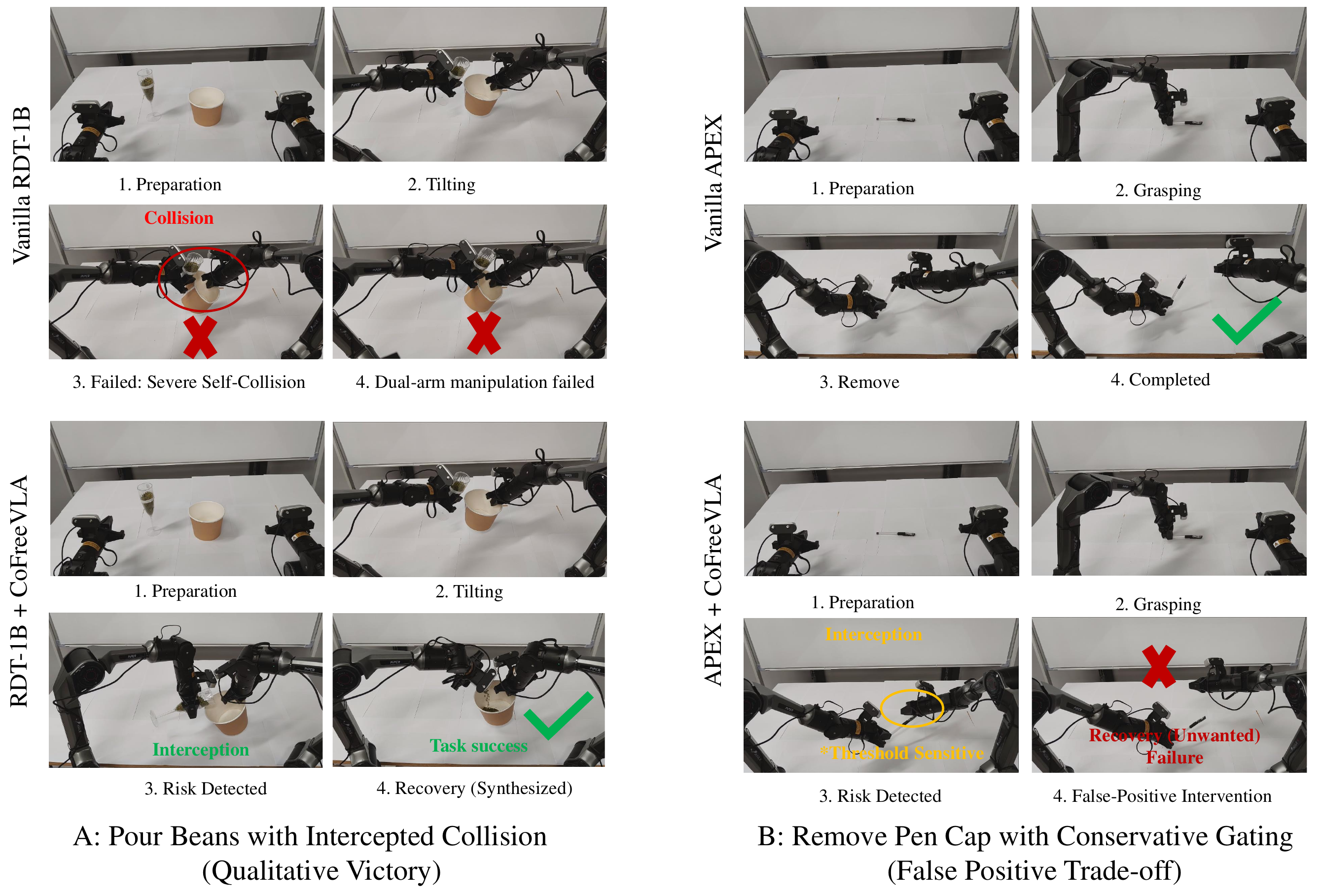}
    \caption{Qualitative comparison of VLA execution with and without the CoFreeVLA framework. (A) Pouring Beans with collision interception and synthesized recovery. (B) Remove Pen Cap with a conservative-gating false-positive intervention.}
    \label{fig:qualitative_comparison}
\vspace{-4mm}
\end{figure*}

Figure~\ref{fig:qualitative_comparison} illustrates how the risk gate modifies closed-loop VLA execution. The qualitative comparison contrasts a nominal rollout with a rollout in which CoFreeVLA screens each proposed action and invokes recovery when the predicted risk exceeds the blocking threshold.

In the Pouring Beans sequence (Fig.~\ref{fig:qualitative_comparison}(A)), vanilla RDT-1B completes the preparation stage and begins tilting the cup, after which the two arms and the grasped object converge within the shared workspace. The resulting contact causes the dual-arm episode to fail. With CoFreeVLA, the gate intercepts the high-risk action before physical contact, and the recovery controller generates a separating motion that returns the arms to a lower-risk configuration. The policy then resumes and completes the task. This qualitative behavior is consistent with the quantitative results in Table~\ref{tab:main_results_comprehensive}: for RDT-1B on Pouring Beans, CR decreases from 0.80 to 0.30 and SR increases from 0.28 to 0.43.

The Remove Pen Cap sequence illustrates the principal failure boundary (Fig.~\ref{fig:qualitative_comparison}(B)). Vanilla APEX executes the preparation, grasping, and removal stages successfully. In the CoFreeVLA rollout, however, the gate flags the intentional close-proximity motion as risky. The resulting intervention and recovery are unnecessary for this valid maneuver, so the episode terminates unsuccessfully. This individual false-positive event coexists with the aggregate improvement for APEX in Table~\ref{tab:main_results_comprehensive}, where Pen Cap Removal CR decreases from 0.23 to 0.16 and SR increases from 0.60 to 0.77.

Taken together, these sequences show that CoFreeVLA is most effective when a nominal plan produces a genuine inter-arm or arm--object overlap. They also show that a fixed conservative threshold can suppress valid actions in millimeter-scale alignment tasks. Task-conditioned calibration is therefore needed to preserve collision reduction without compromising legitimate close-contact manipulation.

\subsection{Ablation Study}
We conduct a controlled ablation using RDT-1B as the backbone. Each variant is evaluated on the same five tasks with 30 trials per task. The configurations form an incremental sequence, followed by targeted removals of the auxiliary geometric heads and hysteresis. We report Avg. CR, Avg. SR, the fraction of control steps blocked by the gate, and the fraction of blocked episodes that return to safe execution and complete the task.

Table~\ref{tab:ablation} summarizes the observed ablation results. The gate alone reduces collisions but leaves blocked episodes without a mechanism for returning to a feasible trajectory. Adding recovery further reduces Avg. CR and increases Avg. SR. Risk-aware policy refinement provides an additional gain by shifting proposed actions away from high-risk plans. Restoring the clearance and time-to-collision heads improves recovery success by providing information about correction direction and urgency. Removing hysteresis increases the blocked-step fraction and slightly reduces SR, consistent with oscillatory switching near the decision boundary. The complete system achieves the best joint safety and task-performance profile. These results confirm that each CoFreeVLA component contributes to lower collision rates and higher task success, with the full configuration providing the strongest overall performance.
\begin{table}[ht]
\centering
\caption{Ablation results on RDT-1B.}
\label{tab:ablation}
\tiny
\setlength{\tabcolsep}{2pt}
\begin{tabular}{lcccc}
\toprule
Variant & Avg. CR $\downarrow$ & Avg. SR $\uparrow$ & Blocked steps $\downarrow$ & Recovery success $\uparrow$ \\
\midrule
Vanilla RDT-1B & 0.62 & 0.40 & -- & -- \\
Gate only & 0.42 & 0.46 & 0.17 & 0.31 \\
Gate + recovery & 0.34 & 0.50 & 0.14 & 0.68 \\
Gate + recovery + refinement$^{\ast}$ & 0.29 & 0.53 & 0.13 & 0.70 \\
Full, no hysteresis & 0.27 & 0.54 & 0.22 & 0.73 \\
\textbf{Full CoFreeVLA} & \textbf{0.25} & \textbf{0.56} & \textbf{0.10} & \textbf{0.82} \\
\bottomrule
\end{tabular}
\begin{tablenotes}[flushleft]
\scriptsize
\item[$^{\ast}$] The refinement variant omits the auxiliary clearance and time-to-collision heads.
\end{tablenotes}
\end{table}

\subsection{Real-Time Control Viability}
We also evaluate whether the safety module can operate within the control budget. The VLA updates at 10\,Hz, while the low-level controller runs at 30\,Hz. The estimator is exported to ONNX or TorchScript, and visual features are prefetched from the VLA backbone. A single estimator forward pass remains below 5\,ms, corresponding to less than 5\% of the 100\,ms VLA update interval and approximately 15\% of one 30\,Hz controller period. The full model therefore leaves sufficient time for candidate evaluation and command dispatch without changing the backbone update rate.

\section{Conclusion} \label{Sec: conclusion}
We present CoFreeVLA, a closed-loop execution layer for screening short-horizon VLA action proposals against dual-arm self-collision risk. The framework combines multimodal risk estimation, command gating, recovery synthesis, and optional action refinement. Across the evaluated bimanual tasks and policy backbones, it consistently reduces collision exposure while improving task completion. Ablation results show that the individual components contribute to the overall safety--performance trade-off, and the estimator operates within the underlying VLA control budget. Fixed conservative thresholds can nevertheless intercept valid close-contact motions in millimeter-scale precision tasks. Future work will investigate task-conditioned thresholds, uncertainty-aware calibration, and adaptive prediction horizons.

\bibliography{references}

\end{document}